\documentclass[conference]{IEEEtran}
\IEEEoverridecommandlockouts

\usepackage{cite}
\usepackage{amsmath,amssymb,amsfonts}
\usepackage{graphicx}
\usepackage{textcomp}
\usepackage{xcolor}
\usepackage{booktabs}
\usepackage{stfloats}
\usepackage{pifont}

\usepackage{rotating}
\usepackage{multirow}
\usepackage{hyperref}
\usepackage{cleveref}

\def\BibTeX{{\rm B\kern-.05em{\sc i\kern-.025em b}\kern-.08em
    T\kern-.1667em\lower.7ex\hbox{E}\kern-.125emX}}

\begin{document}

\title{When Do Foundation Models Pay Off?\\A Break-Even Analysis of Pretrained Time Series Forecasters}

\author{\IEEEauthorblockN{Nicholas Tan Jerome and Frank Simon}
\IEEEauthorblockA{Karlsruhe Institute of Technology (KIT)\\
Eggenstein-Leopoldshafen, Germany\\
\{nicholas.jerome, frank.simon\}@kit.edu}}

\maketitle

\begin{abstract}
Deploying a time series foundation model (FM) requires GPU infrastructure and engineering overhead, with no guarantee of improvement over classical forecasting methods. We provide the first systematic break-even analysis of when this investment pays off. Across 30 benchmark datasets, we compare zero-shot and LoRA fine-tuned FMs (Chronos-Bolt, Moirai, Lag-Llama) against classical baselines (Naive, ETS, ARIMA, XGBoost) at six training set sizes from 2\% to 100\% of available data. FMs outperform classical methods at every evaluated training fraction on 16 of 30 datasets. On 6 datasets, classical methods surpass zero-shot FMs with at most 5\% of the training data (21--6,324 samples); on the remaining 8, break-even ranges from 24 to 22,177 samples. No classical method dominates: ARIMA, Naive, ETS and XGBoost are each best on 6--11 datasets. One robust rule requires no model training: if $n_{\text{train}} < 700$ and seasonality is non-negligible, FM zero-shot is the right low-data choice---below 20\% of the history, classical methods either lose to it (4 of these 10 datasets) or are too short to fit at all (6)---and it stays best at full history on 6 of them; fine-tuning should be skipped. Contrary to common practice, LoRA fine-tuning rarely helps and degrades short series by up to 52\%. We operationalise these findings as a two-step decision framework---compute dataset length and seasonality strength, run a brief 5--10\% pilot only if needed. Cheap dataset features do not predict the break-even category at this benchmark scale, so outside the rule a pilot remains necessary. Code and benchmark: \url{https://github.com/nicolaisi/fm_breakeven}.
\end{abstract}

\begin{IEEEkeywords}
time series forecasting, foundation models, model selection, scaling laws, pretrained models
\end{IEEEkeywords}

\section{Introduction}

Foundation models---large networks pretrained on diverse time series corpora---promise accurate forecasts without task-specific training. Chronos \cite{ansari2024chronos}, Moirai \cite{woo2024moirai}, and Lag-Llama \cite{rasul2023lagllama} have shown impressive zero-shot performance, and the Chronos family alone has tens of millions of downloads on Hugging Face.

Yet practitioners face a fundamental question: \textit{Should I use a foundation model, or stick with a classical method?} Foundation models require GPU infrastructure, have opaque failure modes, and may underperform on domain-specific data; classical methods are interpretable, run on CPUs, and have decades of empirical validation. The cost of a wrong choice is asymmetric: an unnecessary FM means GPU infrastructure, latency, and engineering complexity; sticking with a classical method when an FM would dominate leaves accuracy on the table.

We hypothesize that this comparison is \textit{data-dependent}. With limited task data, FMs may win through pretrained representations; with abundant data, classical methods may catch up by learning task-specific patterns. The \textit{break-even point}---the task data volume at which classical methods begin outperforming FMs---likely varies with intrinsic dataset properties. We address three questions:

\textbf{RQ1: Break-Even Existence and Variation.} Do FMs consistently outperform classical methods, or does the comparison depend on data volume? If break-even points exist, how much do they vary across datasets?

\textbf{RQ2: Determinants of Break-Even.} What dataset properties influence whether and when FMs become beneficial?

\textbf{RQ3: Practical Decision Support.} Are there simple, empirically robust rules practitioners can apply without running experiments, and where are the limits of such prediction at our benchmark scale?

We make three contributions:

\textbf{1. Systematic Benchmark.} A head-to-head comparison across 30 datasets, 6 FM variants (evaluated under up to 3 settings each: zero-shot, LoRA fine-tuning, and full fine-tuning), 4 classical baselines, 6 training size fractions, and 3 seeds---totaling 8,100 runs ($30$ datasets $\times$ $3$ seeds $\times$ [$4$ classical $\times$ $6$ fractions $+$ $6$ zero-shot FMs $+$ $5$ LoRA and $5$ fully fine-tuned FMs $\times$ $6$ fractions]; zero-shot runs once per dataset because it does not depend on the training fraction).

\textbf{2. Break-Even Analysis.} A framework for FM evaluation identifying three dataset categories: FM-dominant (16 of 30), early break-even (classical wins with at most 5\% of data; 6 datasets), and late break-even (classical needs 20\% or more; 8).

\textbf{3. A Robust Decision Rule and a Negative Result on Feature-Based Prediction.} Datasets with $n_{\text{train}} < 700$ and non-negligible seasonality ($S \geq 0.05$) leave FM zero-shot as the only winning or viable choice when only part of the history is available ($<$20\%); as history accumulates, classical methods overtake on half of them. Beyond this rule, four cheap features---series length, seasonal strength, noise ratio, and mean autocorrelation---do not predict the category: a random forest reaches 40.0\% LOO accuracy, below the 53.3\% majority-class baseline. At this benchmark scale, break-even must be measured, not predicted.

\section{Related Work}

\subsection{Time Series Foundation Models}

Chronos \cite{ansari2024chronos} reframed forecasting as language modeling with T5-based models; Chronos-Bolt improved efficiency via direct multi-step patch forecasting. Lag-Llama \cite{rasul2023lagllama} adopted a decoder-only architecture with lag features; Moirai \cite{woo2024moirai} handles varying frequencies and prediction lengths. These models are typically evaluated at fixed data sizes; we examine how their advantage evolves with task data volume.

\subsection{Scaling Laws for Time Series}

Kaplan et al.\ \cite{kaplan2020scaling} established power-law scaling for language models. For time series, Zeng et al.\ \cite{zeng2023transformers} showed that simple linear models can outperform transformers on standard benchmarks; Shi et al.\ \cite{shi2024scaling} and Edwards et al.\ \cite{edwards2024scaling} studied scaling within neural architectures, and Zhou et al.\ \cite{zhou2024fits} showed that models with 10k parameters can match FMs on specific tasks. We instead study scaling \textit{across} paradigms---pretrained FMs versus classical methods that learn only from task data.

\subsection{Meta-Learning for Model Selection}

FFORMS \cite{talagala2023fforms} selects forecasting methods from time series features; FFORMA \cite{montero2020fforma} extended this to model averaging with gradient-boosted meta-learners, achieving top performance in M4 \cite{makridakis2020m4}. Recent work explores online \cite{jakobs2024aalf} and LLM-based selection \cite{wang2025llmselect}. Whereas these predict which method is best at full data, we ask at what training volume the ranking changes---a \textit{threshold estimation} problem prior work does not address.

\subsection{Foundation Model Evaluation}

Benchmarks typically report aggregate metrics at full data \cite{godahewa2021monash}. Xu et al.\ \cite{xu2025specialized} found that properly tuned supervised baselines remain competitive with specialized FMs; Ye et al.\ \cite{ye2024survey} survey evaluation methodology and open challenges. We evaluate across 30 datasets spanning diverse domains and frequencies---substantially broader than the 7-dataset ETT-family benchmarks of prior comparisons---with carefully configured baselines, varying data volumes, and gradient boosting, a strong method often omitted from neural benchmarks.

\section{Methodology}

\subsection{Problem Formulation}

Consider a forecasting task with training data $\mathcal{D}_n = \{(x_t, y_{t+h})\}_{t=1}^{n}$ where $x_t$ represents historical observations and $y_{t+h}$ is the target $h$ steps ahead. Let $f_{\text{FM}}$ denote a foundation model (potentially fine-tuned on $\mathcal{D}_n$) and $f_{\text{CL}}$ denote a classical method trained on $\mathcal{D}_n$.

We define the \textit{break-even point} $n^*$ as the smallest training set size where the classical method matches or exceeds the best foundation model:
\begin{equation}
n^* = \min \{ n : \text{Error}(f_{\text{CL}}, n) \leq \text{Error}(f_{\text{FM}}) \}
\end{equation}
where $f_{\text{FM}}$ is evaluated in zero-shot mode. If no such $n$ exists within the available data range, the classical method never catches up and the dataset is FM-dominant (no finite $n^*$; a dash in Table~\ref{tab:breakeven}). In practice, we observe three patterns: FM-dominant, early break-even ($n^*$ at 2--5\% of data), and late break-even (at 20--100\% of data).

\subsection{Scaling Curve Framework}

We characterize the empirical relationship between training data volume and forecast error by evaluating both model families at discrete training fractions and observing how error changes with $n$.

Foundation models and classical methods exhibit characteristically different scaling behavior in our experiments:

\textbf{Foundation models} produce constant error as a function of $n$ in zero-shot mode---performance is independent of task data by construction. When fine-tuned via LoRA, error may decrease with $n$ on datasets with clear periodicity, or increase due to overfitting on short series.

\textbf{Classical methods} start with higher error at low $n$ but generally improve as more task-specific data becomes available. Their asymptotic performance depends on how well the dataset's structure can be captured by the model family.

The break-even point $n^*$ is where the classical error curve falls below the FM zero-shot baseline. If it never does within the evaluated range, no finite $n^*$ exists and the dataset is FM-dominant. A classical configuration needs at least $\max(m, 15)$ training samples (Sec.~\ref{sec:protocol}), so on short datasets the grid starts above 2\%; when a classical method already wins at the first fraction it can be fitted on (10 of the 14 datasets with a break-even), $n^*$ is an upper bound. We determine break-even empirically from the discrete fraction grid rather than by fitting parametric curves, which avoids extrapolation artefacts on small datasets. We partition datasets into three categories based on the smallest training fraction at which any classical method first matches or beats FM zero-shot: \textit{FM-dominant} (no break-even observed), \textit{Early} ($\leq$5\% of data), and \textit{Late} ($\geq$10\%; all observed Late cases are at $\geq$20\%).

\subsection{Dataset Selection}
\label{sec:datasets}

We select 30 benchmark datasets spanning diverse domains, frequencies, and characteristics. Table~\ref{tab:datasets} summarizes dataset properties.

\begin{table}[t]
\centering
\caption{Benchmark Dataset Characteristics}
\label{tab:datasets}
\scriptsize
\setlength{\tabcolsep}{2.5pt}
\resizebox{\columnwidth}{!}{%
\begin{tabular}{@{}lllrrr|lllrrr@{}}
\toprule
Dataset & Dom. & Fr. & Len. & V & $h$ & Dataset & Dom. & Fr. & Len. & V & $h$ \\
\midrule
Dominick     & Retail & 1W & 90   & 1 & 8  & Exchange     & Fin.    & 1D    & 7,588   & 8   & 30 \\
Hospital     & Health & ME & 84   & 1 & 12 & KDD-Cup-H.   & Air     & 1H    & 10,898  & 1   & 48 \\
M4-Mon.-M.   & Mixed  & ME & 78   & 1 & 18 & ETTh1        & Energy  & 1H    & 17,420  & 7   & 24 \\
M3-Monthly   & Mixed  & ME & 83   & 1 & 18 & ETTh2        & Energy  & 1H    & 17,420  & 7   & 24 \\
M3-Quarterly & Mixed  & QE & 44   & 1 & 8  & Traffic      & Transp. & 1H    & 17,544  & 862 & 24 \\
Tourism-M.   & Tour.  & ME & 243  & 1 & 24 & Pedestrian   & Transp. & 1H    & 22,259  & 1   & 24 \\
COVID-Deaths & Health & 1D & 212  & 1 & 30 & Saugeen      & Hydro.  & 1D    & 23,741  & 1   & 30 \\
M4-Daily     & Mixed  & 1D & 379  & 1 & 14 & Electricity  & Util.   & 1H    & 26,304  & 321 & 24 \\
M4-Weekly    & Mixed  & 1W & 379  & 1 & 13 & Weather      & Climate & 10min & 52,696  & 21  & 720 \\
Rideshare    & Transp.& 1H & 329  & 1 & 24 & Solar-10min  & Energy  & 10min & 52,560  & 1   & 60 \\
ILI          & Health & 1W & 966  & 7 & 24 & ETTm1        & Energy  & 15min & 69,680  & 7   & 96 \\
M4-Hourly    & Mixed  & 1H & 700  & 1 & 24 & ETTm2        & Energy  & 15min & 69,680  & 7   & 96 \\
FRED-MD      & Econ.  & ME & 728  & 1 & 12 & Sunspot      & Science & 1D    & 73,924  & 1   & 30 \\
NN5-Daily    & Fin.   & 1D & 791  & 1 & 56 & AusElec.     & Util.   & 30min & 230,736 & 1   & 48 \\
USBirths     & Health & 1D & 7,305 & 1 & 30 & Wind-Farms  & Energy  & 1min  & 527,040 & 1   & 60 \\
\bottomrule
\end{tabular}}
\\[0.3em]
{\footnotesize Univariate forecasting (OT target) throughout; V = variables, $h$ = horizon. Lengths after multi-series aggregation (Sec.~\ref{sec:datasets}).}
\end{table}

Selection criteria include: (1) public availability for reproducibility, (2) domain diversity spanning energy, transport, finance, health, and climate, (3) sufficient length to enable meaningful subsampling, and (4) prior use in forecasting benchmarks. Datasets that contain many series (the Monash collections and M4) are reduced to a single representative series: we take up to $N$ series spaced evenly through the collection ($N=50$; 100 for M4-Monthly, 200 for M4-Weekly), drop series shorter than the 25th percentile of their lengths, truncate the rest to their most recent 25th-percentile length, z-score each series, and average them. Table~\ref{tab:datasets} reports lengths after this aggregation. One dataset required additional preprocessing: the averaged Rideshare series contains about 200 constant hourly values (COVID-era zero padding in the source data). Constant stretches of at least two seasonal periods are removed before the train/test split (329 of 541 points remain); otherwise the smallest training windows are constant, which makes the MASE denominator zero and the classical fits degenerate.

\subsection{Model Selection}
\label{sec:models}

\subsubsection{Foundation Models}

We evaluate six foundation model variants spanning three model families:

\textbf{Chronos family} \cite{ansari2024chronos}: Chronos-Bolt Small (48M parameters) and Chronos-Bolt Base (205M). These use a patch-based T5 architecture with direct multi-step forecasting and a built-in quantile loss for supervised fine-tuning.

\textbf{Moirai family} \cite{woo2024moirai}: Moirai-Small (14M), Moirai-Base (91M), and Moirai-Large (311M). These use an encoder-only transformer with mixture distribution outputs; we fine-tune them with a negative log-likelihood loss on the prediction patches in our own training loop (the MoiraiFinetune interface is not used).

\textbf{Lag-Llama} \cite{rasul2023lagllama}: A 2.4M parameter decoder-only model using lag features for probabilistic forecasting. Evaluated in zero-shot mode only, as its GluonTS-based training loop does not expose a compatible interface for LoRA adapter injection without substantial modification to the checkpoint loading pipeline.

For Chronos and Moirai, we evaluate three settings: \textit{zero-shot} (direct inference without task-specific training), \textit{full fine-tuning} (updating all pretrained weights on task data), and \textit{LoRA fine-tuning} \cite{hu2022lora} (rank-8 adapters, freezing backbone weights). Lag-Llama is evaluated zero-shot only. Full fine-tuning results (updating all weights) are collected but omitted from main tables due to instability (Sec.~\ref{sec:results}); LoRA is the reported fine-tuning setting for Chronos and Moirai.

\subsubsection{Classical Baselines}

We include four classical methods representing different paradigms:

\textbf{Naive}: Seasonal naive baseline repeating the last observed seasonal cycle. Despite simplicity, this is often competitive for highly seasonal data.

\textbf{ETS}: Exponential smoothing with automatic model selection \cite{hyndman2008automatic} (statsforecast AutoETS).

\textbf{ARIMA}: Auto-ARIMA with automatic order selection \cite{hyndman2008automatic} (statsforecast AutoARIMA), searching over ARIMA$(p,d,q)$ and seasonal variants. For both ETS and ARIMA, the season length is $\min(m, 24)$ and the model is fitted on the most recent $\min(n, 200)$ training samples, which keeps fitting tractable on long and high-frequency series. At forecast time, the fitted model is applied to the history before the test window (at most 2,000 points) without re-estimating its parameters.

\textbf{XGBoost}: Gradient boosting \cite{chen2016xgboost} on lag, rolling-window, and calendar features, forecasting recursively one step at a time. Features are lags $1,\dots,\min(2m, 48)$; rolling mean and standard deviation over the last $w$ values, $w \in \{\lfloor m/2\rfloor, m, 2m\}$ (at least 2); and calendar indicators given by the position of each time step in the seasonal cycle ($t \bmod m$), plus day of week for sub-daily series and for daily series whose period is not a week. Calendar positions use the time index, because most aggregated datasets carry no real timestamps. On short training sets, lags and windows are capped at $\lfloor n/2 \rfloor$. This represents modern machine learning approaches competitive on tabular data.

\subsection{Experimental Protocol}
\label{sec:protocol}

\subsubsection{Training Size Variation}

For each dataset, we create training sets at six logarithmically-spaced fractions: 2\%, 5\%, 10\%, 20\%, 50\%, and 100\% of available training data. Subsampling preserves temporal order by taking the most recent fraction, mimicking realistic scenarios with limited historical data. A classical configuration is run only if its training set has at least $\max(m, 15)$ samples; smaller configurations are omitted, which affects the lowest fractions of the short datasets. Fine-tuned FMs use the most recent $\max(2m, \lfloor f n \rfloor)$ samples at fraction $f$.

\textbf{Forecast context.} At forecast time, every model conditions on all history before the test window: the training pool, the middle 20\%, and earlier test points. FMs use this history up to their own context limits (Chronos-Bolt 2,048 points, Moirai 512, Lag-Llama its checkpoint's 32-step context plus lag features reaching back up to 1,092 steps; shorter histories are zero-padded and masked). Classical methods are \textit{trained} only on the current fraction but also forecast from this history. Zero-shot FM error is therefore independent of the training fraction, and zero-shot is run once per dataset; FM error curves are horizontal by construction. This mirrors deployments in which the raw history is available at inference time and the open question is how much task data a model must be \textit{trained} on; FM curves therefore measure pretraining plus full-context inference, not pretraining alone. Restricting FM context to the same fraction would isolate the pretraining effect and is left to future work.

\subsubsection{Evaluation Protocol}

All datasets use a chronological 60/20/20 split. The first 60\% is the training pool; training sets are its most recent fraction. Each series is z-scored with the mean and standard deviation of the training pool. The middle 20\% is never used for training or model selection, but is provided as context at forecast time. The final 20\% is the test segment: we forecast up to five evenly spaced, non-overlapping windows of length $h$ within it ($h$ is capped at the test-segment length), each from the history directly before the window, and average the absolute errors over windows. Each configuration is repeated with three seeds (0, 1, 2), which seed NumPy and PyTorch; classical methods are deterministic, so their three seeds give identical results.

\subsubsection{Metrics}

Our primary metric is Mean Absolute Scaled Error (MASE) \cite{hyndman2006another}. \textbf{Important:} to ensure comparability across training fractions, $T$ in the denominator is always fixed to the full training set ($\text{frac}=1.0$) rather than the current fraction; only the numerator (forecast error) varies across fractions. The same fixed denominator (the in-sample seasonal-na\"ive MAE with lag $m$ on the full training pool) is used for all models and settings, including LoRA. The seasonal period $m$ is set per-dataset: $m=24$ for hourly datasets (ETTh1/2, Electricity, Traffic, M4-Hourly), $m=96$ for 15-minute datasets (ETTm1/2), $m=144$ for 10-minute Weather, $m=52$ for weekly ILI, $m=5$ for daily Exchange, and $m=1440$ for minutely Wind-Farms.

\textbf{Seed variance.} All FM results are means over three seeds; the best zero-shot FM of a dataset is the model with the lowest seed mean. For FM zero-shot models at $\text{frac}=1.0$, seed standard deviations of the best model per dataset average 0.004 (median 0.000, as Chronos-Bolt is deterministic; maximum 0.024, Moirai-S on M4-Daily). Classical methods are deterministic given fixed data, seed std~$\approx 0$. LoRA fine-tuning seed std averages 0.058 at low fractions ($\leq$5\%, max 0.97) and 0.066 at full data; full fine-tuning is higher at low fractions (mean 0.095), consistent with LoRA being the more stable fine-tuning default in the low-data regime. Full seed-level MASE values are provided in the repository.

\subsubsection{Computational Setup and Reproducibility}

Experiments run on a workstation with an NVIDIA RTX 5090 GPU (32 GB GDDR7 VRAM). Foundation model inference and fine-tuning use the GPU in bfloat16; classical methods use the CPU. Total GPU compute time for all FM experiments was approximately 4 GPU-hours (Chronos/Moirai zero-shot, LoRA and full fine-tuning across 30 datasets; Lag-Llama under a minute); classical experiments took about 1 hour on the CPU in parallel.

\textbf{LoRA configuration.} For Chronos, adapters target \texttt{q} and \texttt{v} projections ($r=8$, $\alpha=16$, dropout=0.05). For Moirai, adapters target \texttt{q\_proj}, \texttt{k\_proj}, \texttt{v\_proj}, \texttt{out\_proj} (same hyperparameters). Both use AdamW ($\text{lr}=5\cdot10^{-4}$ for LoRA and $10^{-4}$ for full fine-tuning, weight decay 0.01, gradient clipping at norm 1), one randomly sampled window per step (context up to 512 points), for $\min(300, \max(50, \lfloor n/h \rfloor))$ steps. Chronos targets are capped at the model's native prediction length (64).

\textbf{XGBoost} features are described in Sec.~\ref{sec:models} (\texttt{n\_estimators}=100, \texttt{max\_depth}=6, \texttt{lr}=0.1).

\textbf{Code and data.} All feature extraction, experiment scripts, and analysis code are released at \url{https://github.com/nicolaisi/fm_breakeven}. Datasets are publicly available from their original sources; download links and preprocessing scripts are included in the repository.

\subsection{Dataset Feature Extraction}

We extract six features capturing intrinsic data properties that may influence break-even behavior. All features are computed from training data without model fitting.

\textbf{ACF decay lag ($\tau_{\text{acf}}$):} The lag at which the autocorrelation function first falls below 0.5, measuring how quickly temporal memory decays. Reported in Table~\ref{tab:breakeven} for reference but capped at 200 for long-memory series, limiting its discriminative power.

\textbf{Mean absolute autocorrelation ($\overline{|\rho|}_{50}$):} The mean of $|\rho(k)|$ for $k = 1, \ldots, 50$, measuring sustained temporal memory without saturation. Used in the classifier as the unsaturated replacement for $\tau_{\text{acf}}$.

\textbf{Hurst exponent ($H$):} Estimated via rescaled-range (R/S) analysis, measuring long-range dependence. $H = 0.5$ indicates no memory; $H > 0.5$ indicates persistence. R/S saturates near 1 for long-memory series, so $H$ is reported for reference only.

\textbf{Noise ratio ($\sigma_r$):} Ratio of differenced to original standard deviation:
\begin{equation}
\sigma_r = \frac{\text{std}(x_t - x_{t-1})}{\text{std}(x_t)}
\end{equation}
Higher values indicate noisier series dominated by innovations.

\textbf{Seasonal strength ($S$):} From STL decomposition:
\begin{equation}
S = 1 - \frac{\text{Var}(R)}{\text{Var}(Y - T)}
\end{equation}
where $R$ is the remainder and $T$ is the trend component.

\textbf{Spectral entropy ($\Phi$):} Entropy of the normalized periodogram, measuring predictability. Low $\Phi$ indicates concentrated spectral power (predictable); high $\Phi$ indicates diffuse power (unpredictable).

Feature extraction completes in under one minute per dataset.

\section{Results}
\label{sec:results}

\subsection{Scaling Curves and Break-Even Points (RQ1)}

\begin{figure*}[!t]
\centering
\includegraphics[width=\textwidth]{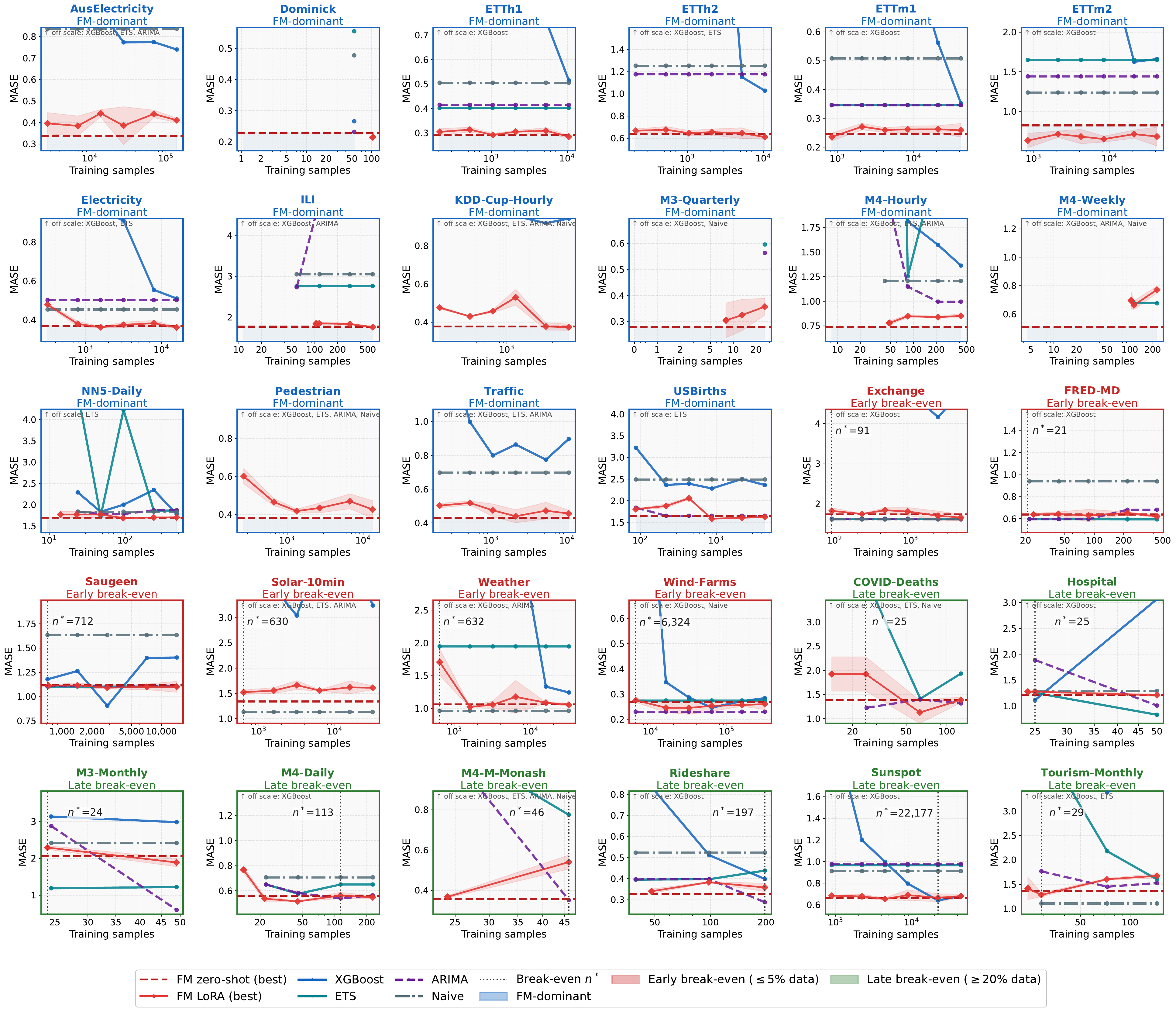}
\caption{Scaling curves across all 30 datasets. \textbf{Dashed horizontal}: best FM zero-shot MASE (mean over 3 seeds). \textbf{Solid red with diamonds}: best FM LoRA MASE (mean over 3 seeds, band $=\pm$1 std). \textbf{Coloured lines}: classical methods. \textbf{Vertical dotted}: break-even $n^*$, ranging from 21 (FRED-MD) to 22,177 samples (Sunspot). Border colour: blue = FM-dominant, red = Early ($\leq$5\%), green = Late ($\geq$20\%).}
\label{fig:scaling_curves}
\end{figure*}

\begin{table*}[!t]
\centering
\caption{Break-Even Points and Dataset Features (30 Datasets)}
\label{tab:breakeven}
\scriptsize
\setlength{\tabcolsep}{3.5pt}
\resizebox{\textwidth}{!}{%
\begin{tabular}{@{}llrlrrrrrrrr@{\hspace{10pt}}llrlrrrrrrrr@{}}
\toprule
Dataset & Best FM & MASE & Best CL & MASE & $n^*$ & $n_{\text{tr}}$ & $\tau_{\text{acf}}$ & $H$ & $\sigma_r$ & $S$ & $\Phi$ &
Dataset & Best FM & MASE & Best CL & MASE & $n^*$ & $n_{\text{tr}}$ & $\tau_{\text{acf}}$ & $H$ & $\sigma_r$ & $S$ & $\Phi$ \\
\midrule
\multicolumn{12}{@{}l}{\textit{FM-dominant} (no break-even)} &
\multicolumn{12}{l}{\textit{Early}} \\
M3-Quarterly & Chr-Bolt-B & 0.279 & ARIMA & 0.564 & --- & 26 & 5 & 1.00 & 0.19 & 0.55 & 0.51 &
FRED-MD & Moirai-L & 0.638 & ETS & 0.594 & 21 & 436 & 81 & 0.99 & 0.11 & 0.04 & 0.27 \\
Dominick & Moirai-B & 0.227 & ARIMA & 0.232 & --- & 54 & 2 & 0.77 & 0.94 & 0.00 & 0.76 &
Exchange & Moirai-B & 1.742 & Naive & 1.615 & 91 & 4,552 & 200 & 1.00 & 0.05 & 0.00 & 0.18 \\
M4-Weekly & Chr-Bolt-S & 0.509 & ETS & 0.676 & --- & 227 & 6 & 0.91 & 0.42 & 0.85 & 0.45 &
Saugeen & Chr-Bolt-S & 1.117 & ETS & 1.105 & 712 & 14,244 & 10 & 0.93 & 0.36 & 0.41 & 0.63 \\
M4-Hourly & Chr-Bolt-B & 0.738 & ARIMA & 0.996 & --- & 420 & 5 & 0.65 & 0.27 & 0.99 & 0.33 &
Solar-10min & Chr-Bolt-B & 1.339 & Naive & 1.133 & 630 & 31,536 & 21 & 0.98 & 0.06 & 0.91 & 0.17 \\
NN5-Daily & Moirai-L & 1.697 & XGBoost & 1.779 & --- & 474 & 1 & 0.55 & 1.11 & 0.88 & 0.52 &
Weather & Moirai-B & 1.061 & Naive & 0.963 & 632 & 31,617 & 200 & 1.00 & 0.04 & 0.75 & 0.31 \\
ILI & Chr-Bolt-B & 1.767 & ETS & 2.760 & --- & 579 & 68 & 0.97 & 0.23 & 0.92 & 0.33 &
Wind-Farms & Chr-Bolt-S & 0.269 & ARIMA & 0.230 & 6,324 & 316,224 & 200 & 1.00 & 0.03 & 0.42 & 0.26 \\
USBirths & Chr-Bolt-B & 1.645 & ARIMA & 1.652 & --- & 4,383 & 2 & 0.67 & 0.85 & 0.84 & 0.37 &
\multicolumn{12}{l}{\textit{Late}} \\
KDD-Cup-Hourly & Chr-Bolt-S & 0.377 & XGBoost & 0.942 & --- & 6,538 & 29 & 0.97 & 0.17 & 0.03 & 0.56 &
M4-Monthly-M. & Moirai-L & 0.357 & ARIMA & 0.351 & 46 & 46 & 4 & 0.88 & 0.50 & 0.76 & 0.59 \\
ETTh1 & Moirai-L & 0.293 & ETS & 0.403 & --- & 10,452 & 200 & 0.94 & 0.12 & 0.28 & 0.27 &
M3-Monthly & Chr-Bolt-B & 2.061 & ARIMA & 0.602 & 24 & 49 & 1 & 0.83 & 1.01 & 0.72 & 0.72 \\
ETTh2 & Chr-Bolt-S & 0.640 & XGBoost & 1.028 & --- & 10,452 & 200 & 0.86 & 0.11 & 0.63 & 0.21 &
Hospital & Chr-Bolt-B & 1.218 & ETS & 0.831 & 25 & 50 & 1 & 0.73 & 1.57 & 0.58 & 0.75 \\
Traffic & Moirai-B & 0.429 & Naive & 0.698 & --- & 10,526 & 2 & 0.66 & 0.74 & 0.63 & 0.61 &
COVID-Deaths & Chr-Bolt-S & 1.381 & ARIMA & 1.316 & 25 & 127 & 22 & 1.00 & 0.03 & 0.00 & 0.35 \\
Pedestrian & Chr-Bolt-B & 0.381 & XGBoost & 1.291 & --- & 13,355 & 4 & 0.65 & 0.41 & 0.94 & 0.27 &
Tourism-Monthly & Chr-Bolt-S & 1.364 & Naive & 1.108 & 29 & 145 & 17 & 0.88 & 0.38 & 0.95 & 0.37 \\
Electricity & Chr-Bolt-B & 0.369 & Naive & 0.454 & --- & 15,782 & 3 & 0.69 & 0.54 & 0.63 & 0.56 &
Rideshare & Moirai-S & 0.326 & ARIMA & 0.288 & 197 & 197 & 4 & 0.92 & 0.55 & 0.05 & 0.68 \\
ETTm2 & Chr-Bolt-B & 0.824 & Naive & 1.239 & --- & 41,808 & 200 & 1.00 & 0.04 & 0.63 & 0.18 &
M4-Daily & Moirai-S & 0.558 & ARIMA & 0.563 & 113 & 227 & 19 & 0.95 & 0.18 & 0.11 & 0.36 \\
ETTm1 & Chr-Bolt-S & 0.246 & ARIMA & 0.346 & --- & 41,808 & 200 & 1.00 & 0.06 & 0.30 & 0.23 &
Sunspot & Chr-Bolt-B & 0.661 & XGBoost & 0.683 & 22,177 & 44,354 & 200 & 0.87 & 0.35 & 0.14 & 0.49 \\
AusElectricity & Chr-Bolt-B & 0.337 & XGBoost & 0.740 & --- & 138,441 & 8 & 0.90 & 0.18 & 0.93 & 0.28 &
 &  &  &  &  &  &  &  &  &  &  &   \\
\bottomrule
\end{tabular}}
\\[0.3em]
\parbox{\textwidth}{\footnotesize Best FM = best zero-shot FM (Chr-Bolt = Chronos-Bolt); FM MASE = mean over 3 seeds, classical deterministic; Best CL and its MASE at full training data. $n^*$ = first training size where classical beats FM zero-shot. $n_{\text{tr}}$ = training samples (60\% split). $\tau_{\text{acf}}$ (capped at 200) and $H$ (R/S) saturate for long-memory series and are shown for reference; the analysis uses $\overline{|\rho|}_{50}$ instead. Fragile assignments: FM-dominant with full-data margin $<$5\% of FM MASE (USBirths 0.4\%, Dominick 2.2\%, NN5-Daily 4.8\%); Late decided only at 100\% with margin $<$3\% (M4-Monthly-M.\ 1.5\%); Rideshare $S = 0.048$. Classical methods need $\max(m, 15)$ samples, so M3-Quarterly and Dominick are FM-dominant on the 100\% fraction alone, and on M3-Monthly, Hospital, Tourism-Monthly, and COVID-Deaths $n^*$ is the first fraction at which classical methods can be fitted.}
\end{table*}

Figure~\ref{fig:scaling_curves} shows scaling curves for all 30 datasets and Table~\ref{tab:breakeven} reports break-even points. We observe three patterns. \textbf{FM-dominant (16 datasets):} the best zero-shot FM beats the best classical method at every evaluated fraction; on three (USBirths, Dominick, NN5-Daily) the full-data margin is below 5\% and could reverse. \textbf{Early break-even (6):} classical methods surpass zero-shot FMs with at most 5\% of the data (21--6,324 samples) on FRED-MD, Exchange, Solar-10min, Weather, Saugeen, and Wind-Farms, and the advantage persists to 100\%. \textbf{Late (8):} classical methods need 20--100\% of training data (24--22,177 samples); with limited data, pretraining provides substantial benefit. M4-Daily and Sunspot are non-monotonic: classical first wins at 50\%, but FM zero-shot is ahead again at full data. No dataset breaks even at 10\%.

\subsection{Model and Setting Analysis}

\begin{table}[t]
\centering
\caption{Performance at Full Training Data (30 Datasets)}
\label{tab:model_comparison}
\scriptsize
\setlength{\tabcolsep}{4pt}
\begin{tabular}{@{}lcccc|lcc@{}}
\toprule
& \multicolumn{2}{c}{Zero-shot} & \multicolumn{2}{c|}{LoRA FT} & & & \\
Model & MASE & $k$ & MASE & $k$ & Classical & MASE & $k$ \\
\midrule
Chronos-Bolt-S & 0.88 & 8  & 1.04 & 8  & ARIMA   & 1.64 & 11 \\
Chronos-Bolt-B & 0.90 & 12 & 1.00 & 10 & Naive   & 1.41 & 7 \\
Moirai-S       & 1.24 & 2  & 1.25 & 2  & ETS     & 1.71 & 6 \\
Moirai-B       & 1.08 & 4  & 1.14 & 4  & XGBoost & 2.43 & 6 \\
Moirai-L       & 1.04 & 4  & 1.12 & 6  & & & \\
Lag-Llama      & 1.53 & 0  & --   & -- & & & \\
\bottomrule
\end{tabular}
\\[0.3em]
{\footnotesize Mean MASE over 30 datasets and 3 seeds; $k$ = number of datasets where the model is best within its setting (zero-shot, LoRA) or among the four classical baselines. Mean MASE is dominated by a few high-MASE datasets (Solar-10min, M3-Monthly, USBirths, ILI, NN5-Daily, Exchange); $k$ is the recommended ranking criterion.}
\end{table}

Table~\ref{tab:model_comparison} compares FM variants and settings.

\textbf{Finding 1: Zero-shot is already competitive.} Zero-shot FMs beat fully-trained classical methods on 18 of 30 datasets: the 16 FM-dominant cases plus M4-Daily and Sunspot. Since $n^*$ uses the \textit{best} FM per dataset, a practitioner committed to a single FM family will typically see lower $n^*$.

\begin{figure}[!t]
\centering
\includegraphics[width=\columnwidth]{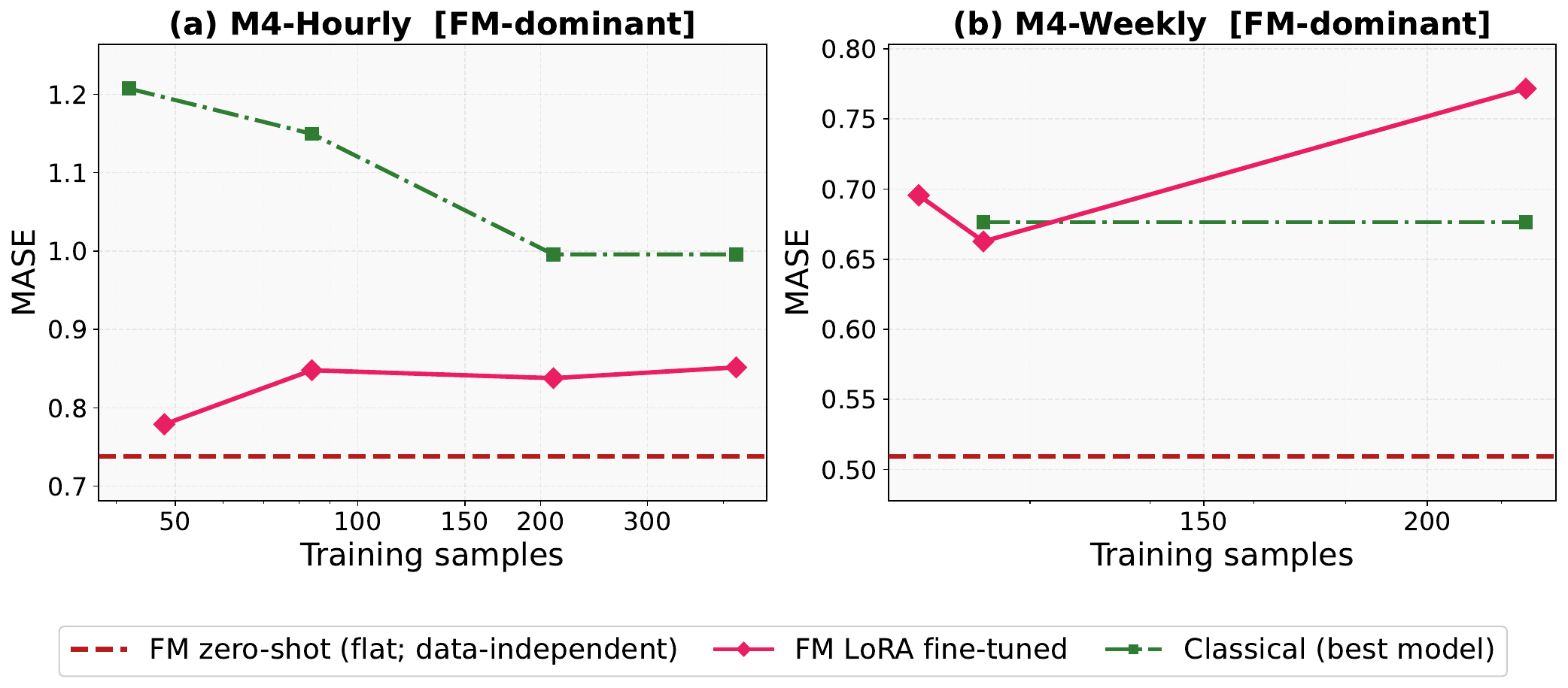}
\caption{Zero-shot vs.\ LoRA on two short FM-dominant datasets. \textbf{(a) M4-Hourly}: best ZS Chr-Bolt-B 0.738 vs.\ best LoRA Chr-Bolt-B 0.852 at full data. \textbf{(b) M4-Weekly}: best ZS Chr-Bolt-S 0.509 vs.\ best LoRA 0.772. LoRA never reaches zero-shot accuracy, and classical never beats FM zero-shot in either.}
\label{fig:finetuning}
\end{figure}

\textbf{Finding 2: LoRA fine-tuning rarely helps.} LoRA improves over zero-shot on 16 of 30 datasets at full data, but the median improvement is only 0.6\%: it is more than 5\% better on 3 datasets and more than 5\% worse on 11. The largest degradations are on short series (Fig.~\ref{fig:finetuning}): the best LoRA model is 52\% worse than the best zero-shot model on M4-Weekly, 51\% on M4-Monthly-M., 28\% on M3-Quarterly, 23\% on Tourism-Monthly, 15\% on M4-Hourly (Chr-Bolt-B, 0.852 vs.\ 0.738), and 10\% on Rideshare. On aggregate, Moirai-S LoRA (1.25) is no better than zero-shot (1.24). When zero-shot already captures the relevant structure, fine-tuning adds noise rather than signal; the assumption that fine-tuning generally helps does not hold unconditionally.

\textbf{Finding 3: Full fine-tuning rarely helps.} At full data it beats LoRA and zero-shot on only 9 of 30 datasets each. At fractions $\leq$5\% it occasionally fails severely: Moirai-L reaches MASE 20.1--26.0 on Hospital versus zero-shot 1.22 (6 runs with MASE $>$ 10). LoRA is more stable in this regime (seed std 0.058 vs.\ 0.095) and is the safer default.

\textbf{Finding 4: No single FM dominates, but Chronos-Bolt leads.} Chronos-Bolt is the best zero-shot FM on 20 of 30 datasets (Base 12, Small 8); Moirai variants win on 10 (Dominick, Traffic, Exchange, Weather, ETTh1, NN5-Daily, FRED-MD, M4-Monthly-M., M4-Daily, Rideshare); Lag-Llama is best on none. Model selection is dataset-dependent even within the FM class.

\textbf{Finding 5: No classical method dominates.} At full data, ARIMA is best on 11 datasets, Naive on 7, and ETS and XGBoost on 6 each. XGBoost wins mainly on long, high-frequency series (AusElectricity, ETTh2, KDD-Cup-Hourly, Pedestrian, Sunspot) and NN5-Daily, but has the worst mean MASE (2.43): fitted on lagged raw levels, tree ensembles cannot extrapolate trending series beyond their training range. FM adoption must therefore be justified against the best of several cheap classical methods, not a single one.

\subsection{Determinants of Break-Even (RQ2)}

Table~\ref{tab:breakeven} shows that $\tau_{\text{acf}}$ saturates at 200 and R/S $H$ near 1.0 for many long-memory series, so these scalar summaries cannot distinguish degrees of long-range dependence. The features that vary most across categories are $n_{\text{train}}$, $S$, $\sigma_r$, and (unsaturated) autocorrelation.
\begin{itemize}
\item \textbf{FM-dominant} datasets (16) are diverse: short seasonal series (M3-Quarterly, M4-Weekly, M4-Hourly, NN5-Daily, ILI), long series with fast ACF decay and moderate-to-high seasonality (Electricity, Traffic, Pedestrian, USBirths, AusElectricity), all four ETT datasets, KDD-Cup-Hourly, and Dominick. On M4-Hourly ($n=420$, $S=0.99$), Chr-Bolt-B zero-shot (0.738) beats ARIMA (0.996) at every fraction. Dominick ($n=54$) is FM-dominant because data scarcity eliminates most classical fractions; its margin is only 2\%.
\item \textbf{Early break-even} datasets (6) are smooth and persistent: five of six have $\sigma_r \leq 0.11$ (Saugeen 0.36), and FRED-MD, Exchange, Weather, and Wind-Farms have $\overline{|\rho|}_{50} \geq 0.81$. On all six, the best classical method is a simple extrapolator (Naive, ETS, or ARIMA). Their lengths range from 436 to 316,224, so length alone does not explain early break-even.
\item \textbf{Late break-even} datasets are predominantly short ($n_{\text{train}} < 500$ for 7 of 8), where classical methods need most available data to accumulate signal. Sunspot ($n=44{,}354$) is the sole long-series exception ($n^*=22{,}177$ at 50\%), and FM zero-shot is ahead again at full data.
\end{itemize}

\subsection{Feature-Based Hypotheses (RQ3)}
\label{sec:rq3}

A random forest (500 trees, depth 4) on the four features $n_{\text{train}}$, $S$, $\sigma_r$, and $\overline{|\rho|}_{50}$, evaluated with LOO cross-validation, reaches 40.0\% accuracy (40.0--43.3\% over 20 forest seeds), below the 53.3\% majority-class baseline of always predicting FM-dominant; it recovers none of the 6 Early and only 2 of 8 Late datasets. At this scale, cheap features do not predict the break-even category. Single-feature correlations are suggestive but not conclusive: $\overline{|\rho|}_{50}$ correlates positively with Early ($r = 0.40$, $p = 0.030$) and negatively with Late ($r = -0.43$, $p = 0.019$), and $\sigma_r$ negatively with Early ($r = -0.37$, $p = 0.045$), but none survives a Bonferroni correction over the 18 feature--category tests. We therefore state the following as hypotheses for larger benchmarks, not findings.

\textbf{Hypothesis 1: Data starvation drives late break-even.}
On short datasets, FM zero-shot wins not because pretrained representations are superior, but because classical methods lack enough samples to fit a reliable model. The FM advantage is passive: it disappears as soon as classical methods accumulate sufficient data. All Late datasets except Sunspot have $n_{\text{train}} < 700$, and of the 14 short datasets only FRED-MD ($S=0.04$) breaks even early. We therefore hypothesize that sufficiently short seasonal series exhibit late break-even or FM dominance regardless of their temporal structure, with $n_{\text{train}} \lesssim 700$ as a rough threshold (not a validated boundary)---chosen to include ILI ($n=579$), the largest short dataset in our benchmark, with a clear gap to the next cluster ($n \geq 4{,}383$). Fine-tuning is counterproductive in this regime: LoRA worsens the best zero-shot MASE by 10--52\% on six short datasets (Finding~2), consistent with overfitting on an already limited sample.

\textbf{Hypothesis 2 (not supported): High innovation variance causes early break-even.}
We expected that series dominated by unpredictable innovations offer pretraining nothing to transfer, so classical methods would catch up from little data. The data point the other way: Early datasets have the lowest noise ratios in the benchmark (median $\sigma_r = 0.05$, vs.\ 0.25 for FM-dominant and 0.44 for Late), and Traffic ($\sigma_r=0.74$) is FM-dominant. An exploratory reading is that smooth, persistent series favour early break-even because simple extrapolation is hard to beat even from a small sample; as this pattern was identified on the same data, it needs confirmation on new datasets.

\textbf{Hypothesis 3: Structural regime determines FM dominance for long series.}
For long series ($n_{\text{train}} > 4000$; 10 FM-dominant, 5 Early, 1 Late), data starvation (H1) is no longer the explanation---classical methods have ample data to fit. The determining factor instead becomes whether the dataset's structure matches what FM pretraining can transfer. Fast ACF decay combined with moderate-to-high seasonality favours FM dominance: Electricity ($\tau=3$, $S=0.63$), Traffic ($\tau=2$, $S=0.63$), Pedestrian ($\tau=4$, $S=0.94$), USBirths ($\tau=2$, $S=0.84$), and AusElectricity ($\tau=8$, $S=0.93$) are all FM-dominant. We tentatively attribute this to short memory providing a stable seasonal signal that FM pretraining transfers well, while the fast decay means classical methods cannot exploit long-range dependence to compensate. The condition is not sharp: Saugeen ($\tau=10$, $S=0.41$) is Early, suggesting the boundary depends on factors the current feature set does not capture.

\begin{figure}[!t]
\centering
\includegraphics[width=0.92\columnwidth]{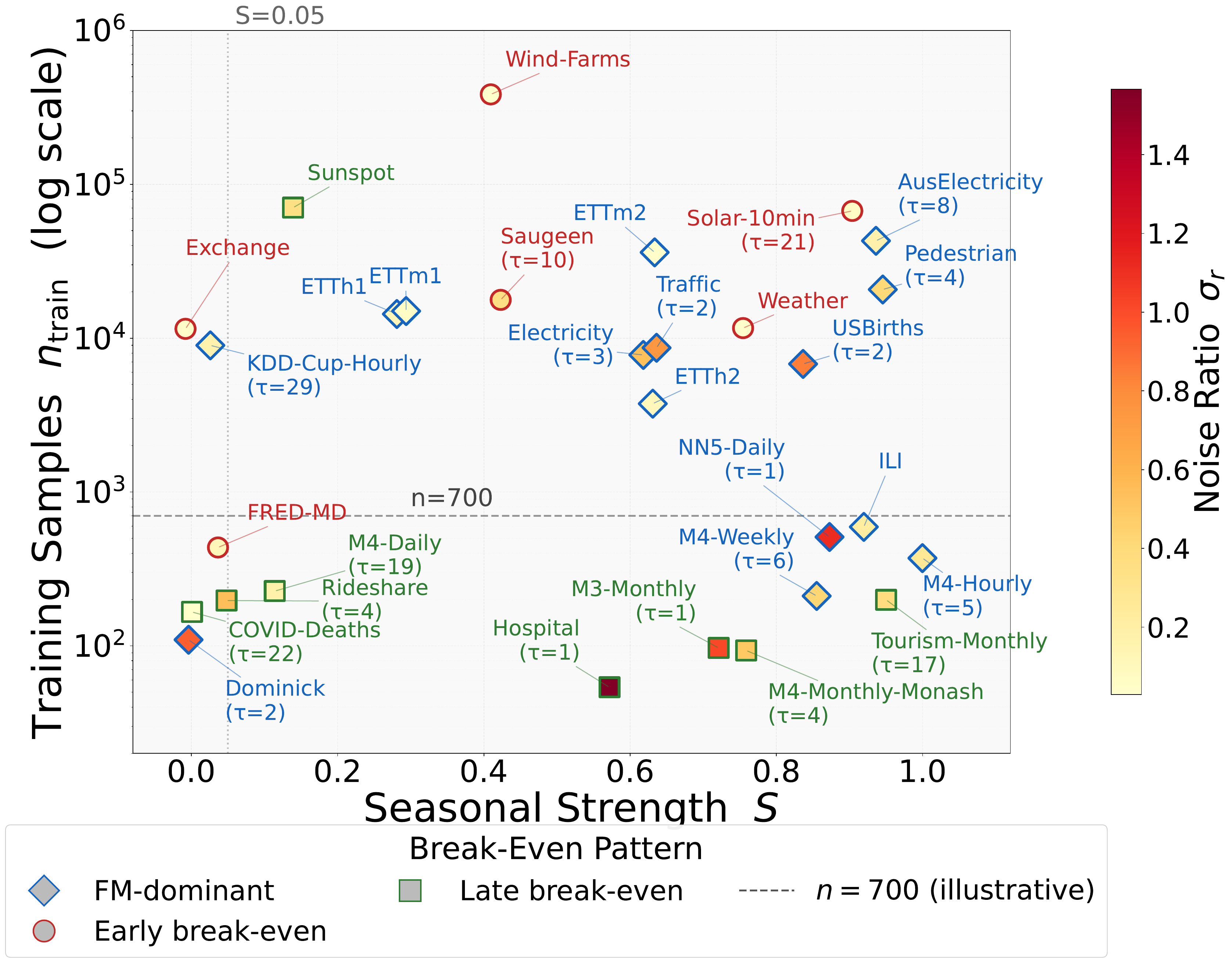}
\caption{Feature space ($n_{\text{train}}$ vs.\ $S$; colour = noise ratio $\sigma_r$; marker = break-even pattern). Dashed lines mark the $n=700$ and $S=0.05$ decision thresholds.}
\label{fig:feature_scatter}
\end{figure}

Figure~\ref{fig:feature_scatter} shows Late datasets clustering at low $n$ regardless of seasonality, consistent with H1. FM-dominant and Early datasets overlap at high $n$ and high $S$ (e.g., Solar-10min Early, $S=0.91$, vs.\ Pedestrian FM-dominant, $S=0.94$); low noise marks most Early datasets but does not resolve the boundary, motivating a pilot-experiment fallback.

\section{Practical Decision Support and Limitations}
\label{sec:rq3practical}

\subsection{Decision Procedure (RQ3)}

Although features do not predict the category in general, one signal provides robust guidance. \textbf{Data starvation rule:} if $n_{\text{train}} < 700$ and $S \geq 0.05$, start with FM zero-shot and avoid LoRA; classical methods cannot accumulate sufficient signal from a fraction of an already short series. On all 10 such datasets, FM zero-shot is unmatched below 20\% of the history: on 4 (M4-Hourly, NN5-Daily, ILI, M4-Daily) classical methods are fitted there and lose, and on the other 6 the series is too short to fit them at all (fewer than $\max(m, 15)$ samples). The decision is final for the 5 FM-dominant cases; for the 5 Late cases it is provisional, since classical methods overtake at 20--100\% of the history (e.g., M3-Monthly: ARIMA 0.602 vs.\ FM 2.061 at full data) and the classical baselines should be re-checked as history accumulates. \textbf{Exception:} if $S < 0.05$ the FM advantage is not guaranteed (FRED-MD: $n^*=21$; COVID-Deaths: $n^*=25$). All other cases require a pilot experiment. Figure~\ref{alg:decision} formalises this as a two-step procedure.

\begin{figure}[!t]
\centering
\fbox{\begin{minipage}{0.95\columnwidth}
\small
\textbf{FM-vs-Classical Deployment Decision}
\vspace{2pt}
\hrule
\vspace{3pt}
\textbf{Input:} $n_{\text{train}}$, seasonal strength $S$\\[2pt]
\textbf{Step 1.} \textit{Data starvation check.} \\
\hspace{1em} If $n_{\text{train}} < 700$ and $S \geq 0.05$:\\
\hspace{2em} $\Rightarrow$ \textbf{Use FM zero-shot. Skip fine-tuning.}\\
\hspace{2em} Re-check vs.\ classical as history grows.\\
\hspace{1em} Otherwise: go to Step 2.\\[2pt]
\textbf{Step 2.} \textit{Pilot experiment.}\\
\hspace{1em} Compare FM zero-shot with the best of Naive,\\
\hspace{1em} ETS, ARIMA, XGBoost trained on 5--10\% of data.\\
\hspace{1em} $\Rightarrow$ \textbf{Winner at pilot fraction guides deployment.}
\vspace{2pt}
\hrule
\end{minipage}}
\caption{Deployment procedure derived from break-even analysis. Step 1 settles the low-data decision on 10 of 30 benchmark datasets (final on 5 FM-dominant, provisional on 5 Late); Step 2 covers the remaining 20.}
\label{alg:decision}
\end{figure}

\subsection{Limitations}

\textbf{Dataset scale:} With 30 datasets, the feature-based classifier (40.0\% LOO) does not beat the majority-class baseline (53.3\%, set by the 16 FM-dominant datasets), and correlations do not survive multiple-testing correction. Testing the hypotheses of Sec.~\ref{sec:rq3} requires 50+ datasets spanning more domains.

\textbf{Baselines:} Classical hyperparameters are not tuned, and XGBoost is fitted on raw levels without differencing, which limits it on trending series; tuned or differenced baselines may move some break-even points earlier.

\textbf{Univariate evaluation:} All experiments target a single output series (OT column). Multivariate FM capabilities---a key advantage of Moirai---are not evaluated.

\textbf{Foundation model evolution:} Our results reflect models available as of early 2026; newer models may shift break-even dynamics.

\textbf{Full fine-tuning} rarely beats zero-shot, fails occasionally at fractions $\leq$5\% (6 runs with MASE~$>$~10, all Moirai-L on Hospital; Finding~3), and is omitted from main tables.

\section{Conclusion}

We introduced break-even analysis as a framework for evaluating foundation models against classical alternatives. Our systematic experiments across 30 datasets reveal that the foundation model advantage is not universal but depends critically on dataset properties and available data volume.

FMs are dominant on 16 of 30 datasets, classical methods (Naive, ETS, or ARIMA) win early on 6 smooth, persistent series, and the remaining 8 break even late.

LoRA fine-tuning rarely helps---it improves the best zero-shot model by more than 5\% on only 3 of 30 datasets---and degrades short series by up to 52\% (M4-Weekly); full fine-tuning beats zero-shot on only 9 of 30. No classical method dominates: ARIMA is best on 11 datasets, Naive on 7, and ETS and XGBoost on 6 each, so FM adoption must be justified against the best of several cheap baselines rather than a single one.

Cheap dataset features do not predict the break-even category at this scale (40.0\% LOO accuracy vs.\ a 53.3\% majority baseline), and our hypothesis that noisy series break even early is not supported. One rule is empirically robust: $n_{\text{train}} < 700$ with $S \geq 0.05$ identifies the low-data FM zero-shot regime on 10 of 30 datasets without any model training (below 20\% of the history, classical methods lose or cannot be fitted)---final on 5, provisional on 5 where classical methods overtake as history accumulates. Beyond this rule, break-even must be measured with a pilot experiment.

We release our benchmark, analysis code, and feature extraction tools at \url{https://github.com/nicolaisi/fm_breakeven}.

\bibliographystyle{IEEEtran}
\bibliography{references}

\end{document}